%% file: main.tex
\documentclass{article} % For LaTeX2e
\usepackage{iclr2027_conference,times}

\input{math_commands.tex}

\usepackage{wrapfig}
\usepackage{graphicx}
\usepackage{hyperref}
\usepackage{url}
\usepackage{algorithm}
\usepackage{algorithmic}
\usepackage{booktabs,multirow,graphicx}

\title{Common-Mode Errors Limit Low-Timestep Deep Spiking Q-Networks}

\author{Zijie Xu, Bingrui Guo, Yiding Sun, Yiting Dong, Zhile Yang, Zhaofei Yu$^\text{†}$  \\
Peking University\\
Beijing, 100871, China \\
}

\iclrfinalcopy 

\begin{document}

\maketitle

\begin{abstract}
Spiking neural networks (SNNs) offer sparse and event-driven computation, making them attractive for energy-constrained reinforcement learning (RL) on edge devices. In value-based RL, deep spiking Q-networks (DSQNs) combine such efficiency with action-value estimation for decision making. However, existing DSQNs often require multiple simulation timesteps for competitive performance, increasing computational and energy costs, whereas reducing the timesteps can cause substantial performance degradation. We investigate this degradation from the perspective of Q-value estimation errors. By decomposing errors across actions into common-mode and differential-mode components, we find that low-timestep DSQNs suffer disproportionately from common-mode errors shared across action values, which are particularly detrimental to temporal-difference learning through bootstrapped targets. Based on this finding, we propose Common-Mode Compensation Deep Spiking Q-Network (CMC-DSQN), which uses an auxiliary ANN to compensate for common-mode errors in the SNN outputs. At inference, greedy action selection can be performed directly from the SNN outputs, allowing the auxiliary ANN to be completely removed and preserving the energy efficiency of SNNs. Extensive experiments on Atari and MiniAtar environments demonstrate substantial performance improvements under low-timestep settings. CMC-DSQN outperforms state-of-the-art DSQN baselines by nearly $20\%$ at $T=2$ and further surpasses the ANN baseline at $T=4$. 

\end{abstract}

\section{Introduction}
\begin{wrapfigure}{r}{0.412\linewidth}
\vspace{-17pt}
\begin{center}
\includegraphics[width=\linewidth]{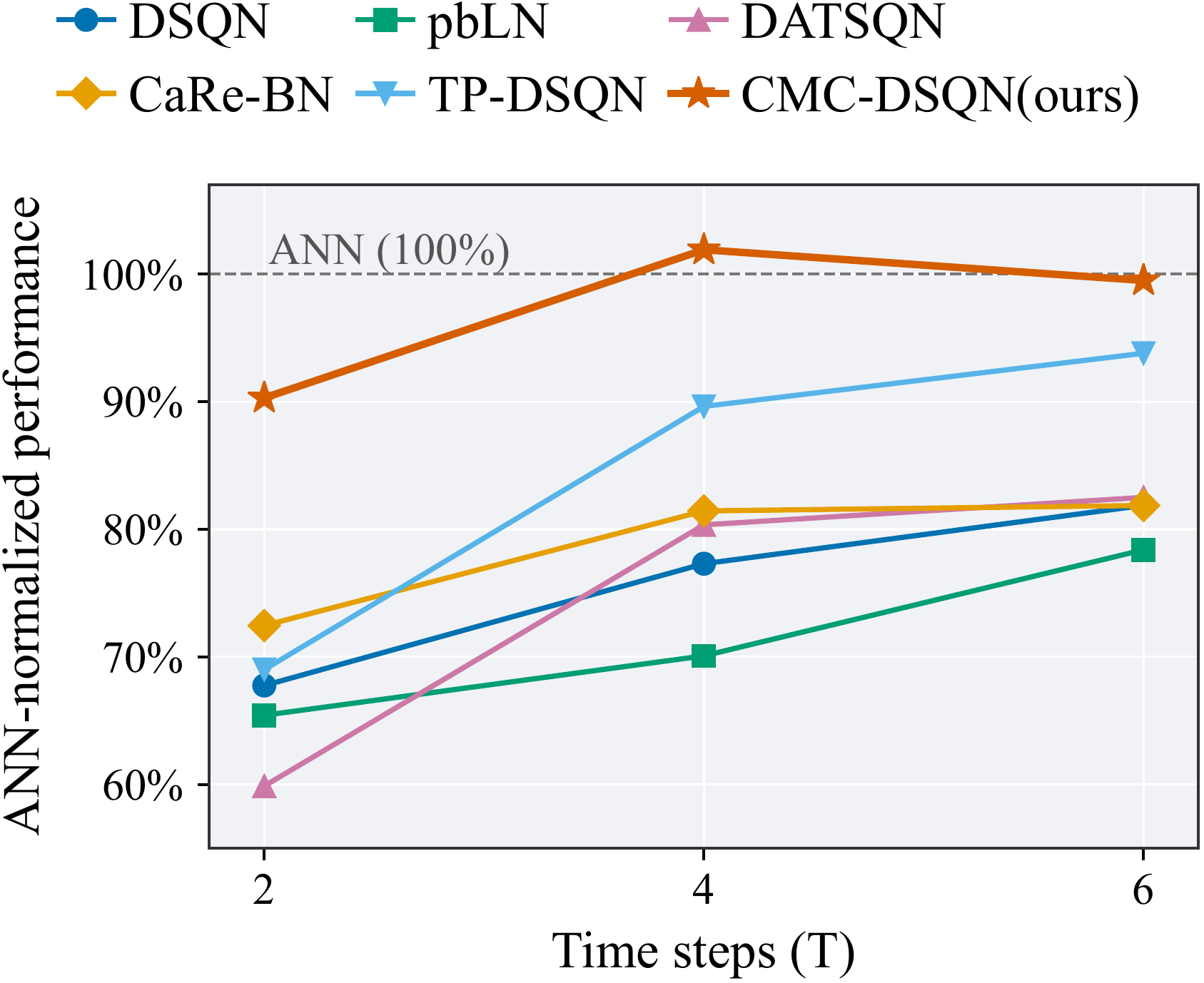}
\vspace{-11.5pt}
\caption{Comparison of existing directly trained DSQNs on the MiniAtar environments~\citep{minatar}, including DSQN~\citep{DQN_BP1}, pbLN~\citep{sun2022solving}, DATSQN~\citep{ghoreishee2026improving}, CaRe-BN~\citep{xu2026care}, and TP-DSQN~\citep{temporal_pruning}.
Performance is averaged across all environments and normalized by the ANN.}
\label{Fig:intro}
\end{center}
\vspace{-25pt}
\end{wrapfigure}
Spiking neural networks (SNNs) process information through sparse binary spikes, providing an event-driven alternative to conventional artificial neural networks (ANNs)~\citep{maass1997networks,gerstner2014neuronal}. When deployed on neuromorphic hardware, such sparse computation can substantially reduce the computational cost and inference latency~\citep{merolla2014million,davies2018loihi,debole2019truenorth}. These properties are particularly attractive for reinforcement learning (RL), where agents repeatedly interact with the environment and continuously make decisions. Combining SNNs with RL therefore provides a promising route toward energy-efficient sequential decision making on resource-constrained edge devices~\citep{SNNadvantage2}.

RL has achieved remarkable success in sequential decision-making tasks~\citep{mnih2015human,PPO,lillicrap2015continuous,haarnoja2018soft}. Among different RL paradigms, value-based methods learn action values that estimate the expected long-term returns of candidate actions. Deep Q-Networks (DQNs) combine Q-learning with ANNs for function approximation~\citep{mnih2013playing,mnih2015human}, achieving human-level control on Atari games and establishing a fundamental framework for value-based RL~\citep{mnih2015human,DoubleDQN,schaul2015prioritized,rainbow}. However, conventional DQNs require dense ANN computation to estimate Q-values at every decision step. Replacing the ANN-based Q-function approximator with an SNN gives rise to Deep Spiking Q-Networks (DSQNs), which aim to improve the energy efficiency of value-based RL.

DSQNs have been developed through both ANN-to-SNN conversion
~\citep{DQN2ANN1,DQN2ANN2,DQN_2ANN_BP} and direct gradient-based training
~\citep{DQN_BP1,BP_DQN_AC,qin2025grsn,sun2022solving,DQN_BP2,ghoreishee2026improving}, achieving competitive performance or even surpassing ANNs. However, such performance often relies on a relatively large number of SNN simulation timesteps. For each RL decision, a DSQN must simulate its neuronal dynamics for $T$ internal timesteps before outputting Q-values. A larger $T$ therefore increases inference latency and computational cost. Although reducing $T$ is highly desirable, we find that low-timestep DSQNs suffer from substantial performance degradation.

In this work, we investigate this degradation from the perspective of Q-value estimation errors. We decompose the errors across actions into \emph{common-mode} and \emph{differential-mode} components. While ANN Q-networks exhibit relatively balanced errors between the two modes, low-timestep DSQNs are dominated by common-mode errors. We further show theoretically that common-mode errors affect temporal-difference learning more directly than differential-mode errors. Experiments on the CliffWalking environment \citep{RL1} confirm their greater impact, showing that removing common-mode errors is more effective than removing differential-mode errors. Together, these results identify common-mode error as a major bottleneck of low-timestep DSQNs.

Motivated by these findings, we propose \emph{Common-Mode Compensation Deep Spiking Q-Network} (CMC-DSQN). To suppress common-mode errors in the low-timestep SNN, CMC-DSQN replaces the action-wise mean of the SNN outputs with a common-mode estimate produced by an auxiliary ANN. The SNN and ANN are jointly optimized through temporal-difference learning, allowing the ANN to provide a more reliable estimate of the common-mode component. Since this common-mode correction does not change the greedy action, the auxiliary ANN is needed only during training and can be removed entirely at inference. CMC-DSQN therefore improves Q-value estimation without introducing additional computational overheads during deployment.

We evaluate CMC-DSQN on both Atari~\citep{mnih2013playing} and
MiniAtar~\citep{minatar} environments. Under low-timestep settings, CMC-DSQN consistently achieves strong performance and substantially outperforms both directly trained DSQNs and ANN-to-SNN conversion baselines. As shown in Fig.~\ref{Fig:intro}, at $T=2$, CMC-DSQN exceeds state-of-the-art DSQN baselines by nearly $20\%$. With $T=4$, CMC-DSQN further surpasses the ANN baseline on both Atari and MiniAtar environments. These results demonstrate that mitigating common-mode estimation errors is particularly effective for closing the performance gap of low-timestep DSQNs.

\section{Related Work}
\subsection{Spiking Neural Networks for Policy-Based Reinforcement Learning}

SNNs have been applied to RL through biologically motivated synaptic plasticity~\citep{R-STDP,R-STDP_3factor,R-STDP_eligibity_trace, continuousAC,SVPG} and gradient-based learning~\citep{learningd_deliema}. For continuous control, numerous studies have developed spiking policies through ANN-to-SNN conversion \citep{xu2026error} or actor--critic learning ~\citep{SDDPG,popSAN,MDC_SAN,ILC_SAN,dynamic_threshold,van2026adaptive,xu2026proxy,xu2026care}. These studies primarily focus on policy learning, whereas our work investigates value-based RL through DSQNs. 

\subsection{Spiking Neural Networks for Value-Based Reinforcement Learning}

For value-based RL, SNNs have been integrated into Deep Q-Networks through
both ANN-to-SNN conversion and direct training. Conversion-based methods
transform pretrained DQNs into spiking networks
~\citep{DQN2ANN1,DQN2ANN2,DQN_2ANN_BP}, whereas directly trained DSQNs use
surrogate gradients such as STBP~\citep{STBP} to optimize spiking Q-functions
end-to-end~\citep{DQN_BP1,BP_DQN_AC}. Subsequent work has improved DSQNs
through recurrent mechanisms~\citep{qin2025grsn}, normalization
~\citep{sun2022solving,xu2026care}, and enhanced spiking-neuron designs
~\citep{DQN_BP2,ghoreishee2026improving}. However, Q-value estimation errors in low-timestep DSQNs remain insufficiently understood.

\subsection{Value Decomposition in Deep Q-Learning}

Value decomposition exploits structures shared across actions. Dueling
architectures and their extensions decompose Q-values into a state-dependent
shared component and action-dependent advantages
~\citep{dueling,rainbow,Branching_Dueling}, while VA-learning further studies
value--advantage learning under bootstrapping and its relation to the behavior
policy~\citep{tang2023va}. 
In contrast to these methods that decompose the Q-function itself, we decompose its estimation errors into common-mode and differential-mode components, and show that common-mode errors dominate in low-timestep DSQNs. Beyond this distinction, CMC-DSQN adopts a hybrid architecture in which a training-only auxiliary ANN estimates the common-mode component without sharing the SNN encoder, differing from typical dueling-style shared-encoder architectures.

\section{Preliminaries}

\subsection{Deep Q-Learning}

Reinforcement learning (RL) considers a Markov decision process (MDP) $\langle\mathcal{S},\mathcal{A},P,R,\gamma\rangle$~\citep{RL1}, where
$\mathcal{S}$ and $\mathcal{A}$ denote the state and action spaces,
$P(s'|s,a)$ is the transition probability, $R(s,a)$ is the reward function,
and $\gamma\in[0,1)$ is the discount factor. At interaction step $i$, the
agent observes state $s_i$, selects action $a_i$, receives reward $r_i$, and
transitions to $s_{i+1}$. The discounted return $G_i$ and the action-value function $Q^\pi(s,a)$
under policy $\pi$ are defined as
\begin{equation}
G_i=\sum_{k=0}^{\infty}\gamma^k r_{i+k},
\end{equation}
\begin{equation}
Q^\pi(s,a)=\mathbb{E}_{\pi}
\left[G_i\mid s_i=s,a_i=a\right].
\end{equation}
The objective is to learn the optimal action-value function $Q^*$, which
satisfies the Bellman equation
\begin{equation}
Q^*(s_i,a_i)
=
\mathbb{E}\left[
r_i+\gamma\max_{a'}Q^*(s_{i+1},a')
\right].
\end{equation}

Deep Q-Networks (DQNs)~\citep{mnih2013playing,mnih2015human} approximate $Q^*$ using a neural network
$Q_\theta(s,a)$ parameterized by $\theta$. Given a transition
$(s_i,a_i,r_i,s_{i+1},d_i)$, the network is optimized by minimizing the
temporal-difference (TD) loss
\begin{equation}
\mathcal{L}
=
\mathbb{E}\left[
\left(Q_\theta(s_i,a_i)-y_i\right)^2
\right],
\label{eq:dqn_loss}
\end{equation}
\begin{equation}
y_i=
r_i+\gamma(1-d_i)\max_{a'}Q_\theta(s_{i+1},a'),
\label{eq:dqn_target}
\end{equation}
where $y_i$ is the bootstrapped TD target and $d_i\in\{0,1\}$ indicates
whether the transition terminates the episode. During inference, the agent
follows the greedy policy
$a_i=\arg\max_{a\in\mathcal{A}}Q_\theta(s_i,a)$.
Because the TD target depends on estimated Q-values at the next state,
estimation errors can propagate through successive bootstrapped updates,
which motivates our analysis in Sec.~\ref{Sec:analysis}.

\subsection{Spiking Neural Networks}

Spiking neural networks (SNNs) process information through discrete spikes over
multiple simulation timesteps. We adopt the commonly used leaky
integrate-and-fire (LIF) neuron with hard reset~\citep{LIF}. At SNN timestep
$t$, its dynamics are
\begin{equation}
\tilde{u}^{t}=\lambda u^{t-1}+I^{t},
\qquad
o^{t}=H(\tilde{u}^{t}-v_{\mathrm{th}}),
\qquad
u^{t}=\tilde{u}^{t}(1-o^{t}),
\end{equation}
where $\tilde{u}^{t}$ and $u^{t}$ denote the membrane potentials before and
after reset, $\lambda$ is the membrane decay factor, $I^{t}$ is
the synaptic input, $v_{\mathrm{th}}$ is the firing threshold, and
$H(\cdot)$ denotes the Heaviside function.

A deep spiking Q-network (DSQN) replaces the ANN-based Q-network with an SNN. For each RL observation, the SNN runs for $T$ simulation timesteps to output the action values. A larger $T$ generally improves value representation at higher computational cost, whereas a smaller $T$ makes accurate Q-value estimation more challenging.

\section{Common-Mode Errors in Deep Spiking Q-Networks}

\begin{figure}[t]
\begin{center}
\includegraphics[width=0.99\linewidth]{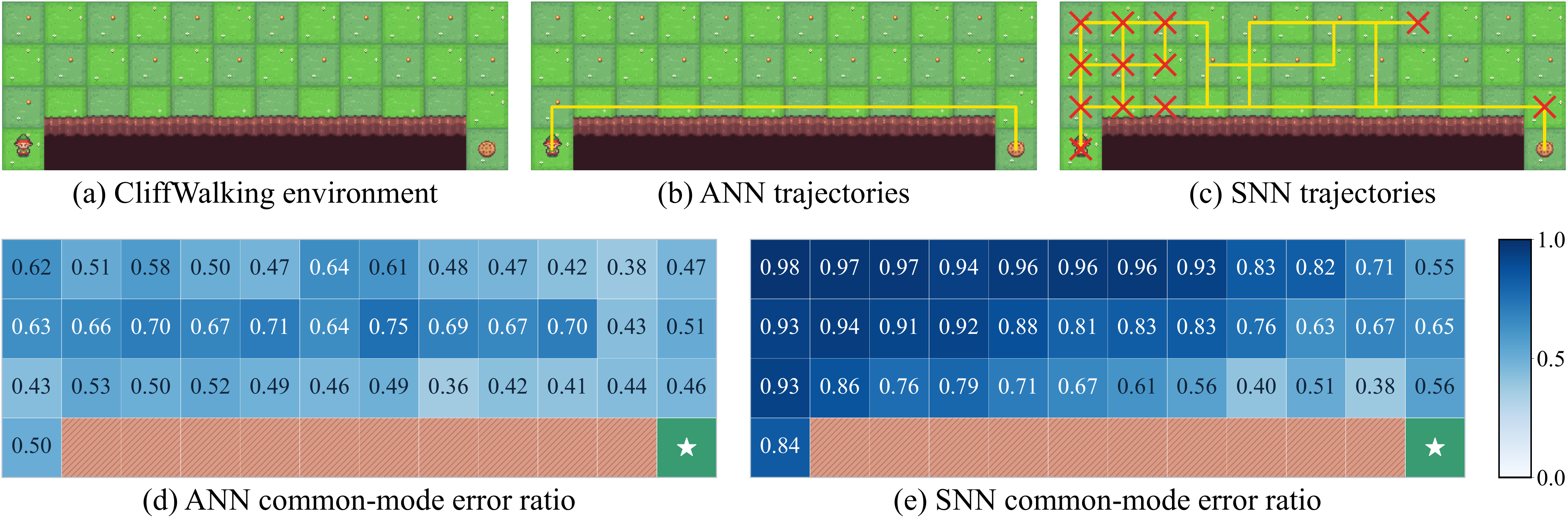}
\caption{
Analysis on deterministic CliffWalking.
(a) Environment.
(b,c) Greedy trajectories of the ANN and low-timestep SNN,
respectively, aggregated over 20 random seeds. Yellow segments denote trajectories, and red crosses mark states leading to infinite loops.
(d,e) State-wise common-mode error ratios of the ANN and SNN with respect to the optimal action-value function $Q^*$.
}
\label{Fig:analysis1}
\end{center}
\end{figure}

We begin with a case study on the deterministic CliffWalking environment
shown in Fig.~\ref{Fig:analysis1}(a), where all transitions are deterministic
and entering a cliff cell terminates the episode. Its small state space allows
both the optimal action-value function $Q^*$ and the action-value function
$Q^\pi$ of any policy to be computed exactly from the Bellman
equations~\citep{Bellman}. This enables direct analysis of Q-value estimation errors and their effects on TD bootstrapping~\citep{TD}.

\label{Sec:analysis}

\subsection{Common- and Differential-Mode Error Decomposition}
Consider a discrete action space $\mathcal{A}$ with $|\mathcal{A}|=K$.
Given a reference action-value function $Q^{\mathrm{ref}}$, we decompose the
estimation error $e(s,a)=\hat{Q}(s,a)-Q^{\mathrm{ref}}(s,a)$ across actions as
\begin{equation}
    e_c(s)
    =
    \frac{1}{K}
    \sum_{a\in\mathcal{A}} e(s,a),
    \label{eq:cm_error}
\end{equation}
\begin{equation}
    e_d(s,a)
    =
    e(s,a)-e_c(s),
    \label{eq:dm_error}
\end{equation}
where $e_c$ is the \emph{common-mode error}, representing the error shared across actions, and $e_d$ is the \emph{differential-mode error},
capturing errors in the relative differences among actions. 

We further define the common-mode error ratio $\rho_c(s)$ to quantify the relative magnitude of the common-mode error compared with
the two error components at each state:
\begin{equation}
    \rho_c(s)
    =
    \frac{E_c(s)}
    {E_c(s)+E_d(s)},\qquad E_c(s)=|e_c(s)|,\qquad
    E_d(s)=
    \sqrt{\frac{1}{K}\sum_{a}e_d(s,a)^2}.
    \label{eq:cm_ratio}
\end{equation}

\subsection{Common-Mode Error Dominance in Low-Timestep SNNs}
\begin{figure}[t]
\begin{center}
\includegraphics[width=0.99\linewidth]{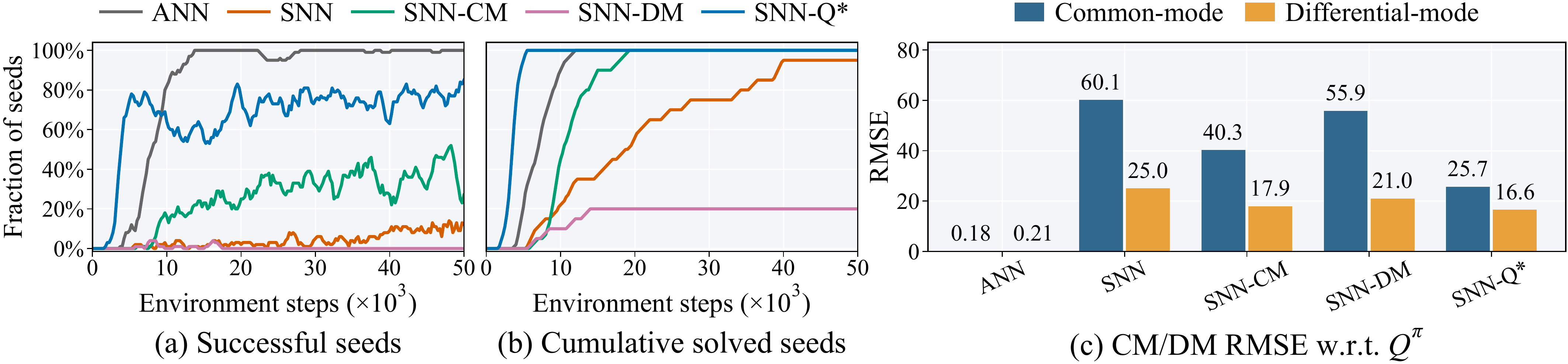}
\caption{
Learning dynamics and Q-value estimation errors on CliffWalking environment.
(a) Fraction of successful seeds at each evaluation point.
(b) Cumulative fraction of seeds that have solved the task at least once.
Curves in (a) and (b) are smoothed for visual clarity.
(c) Common- and differential-mode RMSEs with respect to the exact $Q^\pi$
of each learned deterministic greedy policy.
}
\label{Fig:analysis2}
\end{center}
\end{figure}

We first compare a conventional ANN with a low-timestep SNN ($T=2$) under the
same DQN framework over 20 random seeds. Because both the environment and the
greedy evaluation policy are deterministic, each seed either succeeds or fails
at each evaluation point. As shown in Figs.~\ref{Fig:analysis2}(a) and (b),
the ANN rapidly learns a stable solution, whereas the SNN is substantially less
stable. Although many SNN seeds eventually discover a successful policy, only
a small fraction remain successful at any given evaluation point, indicating
difficulty in maintaining stable value estimates.

\textbf{Common-mode error dominance.}
To separate estimation error from policy suboptimality,
Fig.~\ref{Fig:analysis2}(c) measures the common- and differential-mode RMSEs
with respect to the exact $Q^\pi$ of each learned greedy policy. The SNN
exhibits substantially larger errors than the ANN in both modes. More
importantly, its common-mode error is much larger than its differential-mode
error, whereas the two are relatively balanced in the ANN. Thus, low-timestep
SNNs exhibit not only larger Q-value estimation errors, but also a 
common-mode dominance.

\textbf{State-wise common-mode dominance.}
We further examine the spatial distribution of these errors using the optimal action-value function $Q^*$ as the
reference. Figures~\ref{Fig:analysis1}(d) and (e) show the state-wise
common-mode error ratio $\rho_c(s)$ for the ANN and SNN, respectively. The ANN
maintains a relatively balanced error composition, whereas the SNN is dominated
by common-mode errors at nearly every state, particularly in the upper-left
region. This region also coincides with the
cyclic trajectories frequently observed in unsuccessful SNN policies (Fig.~\ref{Fig:analysis1}(c)), suggesting a close association between common-mode error dominance and unstable policy.

\subsection{Error Propagation through Temporal-Difference Bootstrapping}
To determine whether common-mode errors contribute to performance degradation, we next analyze how common-mode and differential-mode errors propagate through TD bootstrapping. Consider the one-step target in Eq.~(\ref{eq:dqn_target}) for
a non-terminal transition ($d_i=0$). Given a reference Q-function $Q^{\mathrm{ref}}$, let $\hat Q(s,a)=Q^{\mathrm{ref}}(s,a)+e_c(s)+e_d(s,a)$. Substituting this decomposition into the TD target gives
\begin{equation}
\left\{
\begin{aligned}
&\hat y_i
=
r_i+\gamma e_c(s_{i+1})
+\gamma\max_a\!\left[
Q^{\mathrm{ref}}(s_{i+1},a)+e_d(s_{i+1},a)
\right],\\
&y_i^{\mathrm{ref}}
=
r_i+\gamma\max_a Q^{\mathrm{ref}}(s_{i+1},a).
\end{aligned}
\right.
\label{eq:td_targets}
\end{equation}
Thus, the error propagated to the bootstrapped target is:
\begin{equation}
\hat y_i-y_i^{\mathrm{ref}}
=
\gamma e_c(s_{i+1})
+\gamma\Delta_d(s_{i+1}).
\label{eq:bootstrap_error}
\end{equation}
\begin{equation}
\Delta_d(s)
=
\max_a\!\left[
Q^{\mathrm{ref}}(s,a)+e_d(s,a)
\right]
-\max_a Q^{\mathrm{ref}}(s,a).
\label{eq:delta_d}
\end{equation}
The common-mode error shifts all action values equally and therefore enters the
bootstrap target directly, whereas the differential-mode error affects it only
through maximization, with
$|\Delta_d(s)|\leq\max_a|e_d(s,a)|$. Common-mode errors thus have a more
direct pathway into successive TD updates.

\paragraph{Removing Bootstrap Errors.}
To isolate the effects of the two modes, we modify only the next-state Q-values used for bootstrapping:
\begin{equation}
\left\{
\begin{aligned}
\hat Q_{\mathrm{CM}}(s,a)
&=\hat Q(s,a)-e_c^\pi(s),\\
\hat Q_{\mathrm{DM}}(s,a)
&=\hat Q(s,a)-e_d^\pi(s,a),\\
\hat Q_{Q^*}(s,a)
&=Q^*(s,a),
\end{aligned}
\right.
\label{eq:error_removal}
\end{equation}
where $e_c^\pi$ and $e_d^\pi$ are computed with respect to the exact $Q^\pi$
of the current learned greedy policy. These variants are denoted as SNN-CM,
SNN-DM, and SNN-Q*, respectively.

\paragraph{Learning dynamics.}
As shown in Figs.~\ref{Fig:analysis2}(a) and (b), replacing the next-state
estimate with exact $Q^*$ substantially improves learning, confirming the
importance of bootstrap Q-value accuracy. Removing only the common-mode error
also yields a clear improvement, whereas removing the differential-mode error
does not improve learning\footnote{One possible
explanation is that $\Delta_d$ can occasionally offset part of the common-mode
bias; removing it may eliminate this incidental cancellation.}. This indicates that common-mode errors play a more critical role than differential-mode errors in destabilizing TD learning in low-timestep SNNs.

\paragraph{Q-value estimation errors.}
The Q-value estimation errors in Fig.~\ref{Fig:analysis2} (c) further support this observation. Using the exact optimal Q-value $Q^*$ achieves the lowest estimation errors. Removing either the common-mode or differential-mode error reduces both error components, but correcting the common mode leads to a relatively larger reduction. Together, these results show that common-mode errors have a substantially greater impact on TD learning than differential-mode errors.

\section{Common-Mode Compensation for Deep Spiking Q-Networks}
\label{Sec:method}
The analysis in Sec.~\ref{Sec:analysis} shows that common-mode estimation errors can directly bias bootstrapped TD targets and degrade learning in low-timestep DSQNs. In the simple CliffWalking environment, such errors can be explicitly removed because exact reference Q-functions are available. In practical RL tasks, however, neither $Q^\pi$ nor $Q^*$ is accessible during training.

A straightforward alternative is to introduce an auxiliary ANN to estimate and compensate for the entire Q-value estimation error. However, this requires the auxiliary network to model both common- and differential-mode components, introducing unnecessary complexity. As shown in Sec.~\ref{Sec:ablation_study}, this design improves over the vanilla DSQN but underperforms CMC-DSQN.

\subsection{CMC-DSQN}

\begin{algorithm}[t]
\caption{CMC-DSQN}
\label{alg:cmc}
\begin{algorithmic}[1]
\STATE \textbf{Training:}
\FOR{each sampled transition $(s_i,a_i,r_i,s_{i+1})$}
    \STATE Compute $Q_{\phi,\theta}^{\mathrm{CMC}}(s_i,\cdot)$ and
    $Q_{\phi,\theta}^{\mathrm{CMC}}(s_{i+1},\cdot)$ using Eq.~(\ref{eq:cmc_q})
    \STATE Construct the TD target $y_i$ using Eq.~(\ref{eq:dqn_target})
    \STATE Update $\theta$ and $\phi$ by minimizing
    $\mathcal{L}=
    (Q_{\phi,\theta}^{\mathrm{CMC}}(s_i,a_i)-y_i)^2$
\ENDFOR
\STATE \textbf{Inference:}
\STATE Compute only $Q_\theta^{\mathrm{SNN}}(s_i,\cdot)$
\STATE Select $a_i=\arg\max_a Q_\theta^{\mathrm{SNN}}(s_i,a)$
\end{algorithmic}
\end{algorithm}
Motivated by the analysis in Sec.~\ref{Sec:analysis}, we propose \emph{Common-Mode Compensation Deep Spiking Q-Network} (CMC-DSQN), which introduces an auxiliary ANN $C_\phi^{\mathrm{ANN}}(s)$ parameterized by $\phi$ to estimate the common-mode component.

Let $Q_\theta^{\mathrm{SNN}}(s,a)$ denote the raw Q-value produced by the SNN
with parameters $\theta$. Its common-mode estimation error $e_c(s)$ with respect to a reference Q-function $Q^{\mathrm{ref}}$ is
\begin{equation}
e_c(s)
=
\frac{1}{K}
\sum_{a\in\mathcal{A}}
\left(
Q_\theta^{\mathrm{SNN}}(s,a)
-
Q^{\mathrm{ref}}(s,a)
\right).
\end{equation}
Directly predicting $e_c(s)$ is undesirable because the SNN evolves
continuously during training, making the residual error itself a moving
target. Instead, the auxiliary ANN directly estimates the corresponding
common-mode Q-value component,
\begin{equation}
C_\phi^{\mathrm{ANN}}(s)
\approx
\frac{1}{K}
\sum_{a\in\mathcal{A}}
Q^{\mathrm{ref}}(s,a).
\end{equation}
Note that $Q^{\mathrm{ref}}$ is unavailable during training. Thus, the auxiliary ANN is learned jointly with the SNN solely through TD learning by constructing the CMC Q-value as
\begin{equation}
Q_{\phi,\theta}^{\mathrm{CMC}}(s,a)
=
Q_\theta^{\mathrm{SNN}}(s,a)
-
\frac{1}{K}
\sum_{a'\in\mathcal{A}}
Q_\theta^{\mathrm{SNN}}(s,a')
+
C_\phi^{\mathrm{ANN}}(s).
\label{eq:cmc_q}
\end{equation}
Thus, CMC-DSQN removes the common component of the SNN outputs and replaces it with the estimate produced by the auxiliary ANN. The ANN and SNN process the observation independently without sharing an encoder, and are jointly optimized through the DQN objective in Eq.~(\ref{eq:dqn_loss}). The overall training and inference procedures of CMC-DSQN are summarized in Algorithm~\ref{alg:cmc}.

An important property of Eq.~(\ref{eq:cmc_q}) is that the auxiliary ANN does not affect greedy action selection:
\begin{equation}
\arg\max_a Q_{\phi,\theta}^{\mathrm{CMC}}(s,a)
=
\arg\max_a Q_\theta^{\mathrm{SNN}}(s,a).
\end{equation}
Therefore, the auxiliary ANN is required only during training and can be removed entirely at inference, where actions are selected directly from the original SNN outputs. CMC-DSQN therefore introduces no additional computational overhead and preserves the energy efficiency of SNNs.

\begin{figure}[t]
\begin{center}
\includegraphics[width=0.99\linewidth]{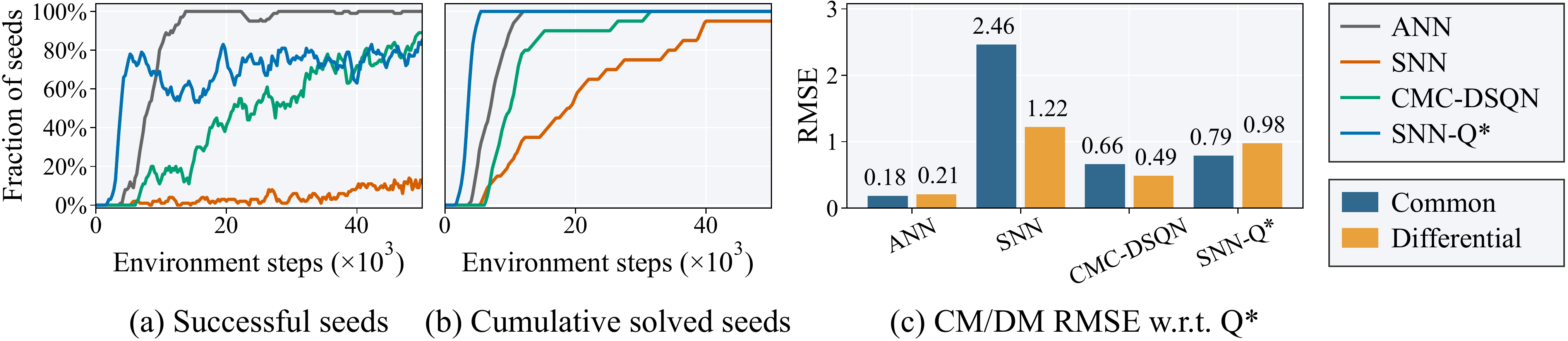}
\caption{
Learning dynamics and Q-value estimation errors on CliffWalking.
(a) Fraction of successful seeds at each evaluation point.
(b) Cumulative fraction of seeds that have solved the task at least once.
Curves in (a) and (b) are smoothed for visual clarity.
(c) Common- and differential-mode RMSEs with respect to the exact optimal
Q-function $Q^*$. Unlike Fig.~\ref{Fig:analysis2}(c), which uses the exact
$Q^\pi$ of each learned policy as the reference, this panel evaluates all
methods against $Q^*$.
}
\label{Fig:analysis3}
\end{center}
\end{figure}

\subsection{Understanding Common-Mode Compensation}
We revisit CliffWalking to examine whether CMC-DSQN alleviates the error characteristics identified in Sec.~\ref{Sec:analysis}. As shown in Figs.~\ref{Fig:analysis3}(a) and (b), CMC-DSQN substantially improves learning stability over the vanilla SNN and enables nearly all seeds to solve the task. Figure~\ref{Fig:analysis3}(c) further shows that CMC-DSQN markedly reduces both common- and differential-mode Q-value errors.

Interestingly, CMC-DSQN achieves even lower estimation errors than SNN-Q*, despite the latter using the exact optimal Q-values for bootstrapping. SNN-Q* still requires a single low-timestep SNN to reduce both common- and differential-mode errors, whereas CMC-DSQN assigns the common-mode error reduction to the ANN and allows the SNN to focus on reducing differential-mode errors. This reduces the representational burden on the SNN and improves estimation of both error modes.
\section{Experiments}
\label{Sec:experiments}

\subsection{Experimental Setup}

We evaluate CMC-DSQN on two commonly used benchmarks: Atari~\citep{mnih2013playing} and MiniAtar~\citep{minatar}. For Atari, we consider seven games evaluated in the original DQN study~\citep{mnih2013playing}: BeamRider, Breakout, Enduro, Pong, Q*bert, Seaquest, and SpaceInvaders. For MiniAtar, we evaluate on all five environments: Asterix, Breakout, Freeway, Seaquest, and SpaceInvaders. We use $T\in\{2,3,4\}$ for Atari and $T\in\{2,4,6\}$ for MiniAtar. All models are trained for 30M environment transitions on Atari and 5M transitions on MiniAtar. Results are averaged over three random seeds for Atari and five random seeds for MiniAtar. The main hyperparameters follow \citet{mnih2015human} and \citet{minatar}, respectively. Detailed implementation settings are provided in the Appendix.

We compare CMC-DSQN with both ANN-to-SNN conversion methods and directly trained DSQNs. The conversion baselines include vanilla ANN-to-SNN conversion with integrate-and-fire (IF) neurons (DSQN-CVT), differential coding with multi-threshold neurons (DCGS)~\citep{huang2025differential,
huang2024towards}, and cross-step residual potential initialization (CRPI)~\citep{xu2026error} combined with DCGS. The directly trained baselines include vanilla DSQN~\citep{DQN_BP1}, potential-based layer normalization (pbLN)~\citep{sun2022solving}, deep asymmetric ternary spiking Q-network (DATSQN)~\citep{ghoreishee2026improving}, confidence-adaptive and re-calibration batch normalization (CaRe-BN)~\citep{xu2026care}, and DSQN with temporal pruning (TP-DSQN)~\citep{temporal_pruning}.

In addition to raw scores, we report the average performance ratio (APR):
\begin{equation}
    \mathrm{APR}_m
    =
    \frac{1}{|\mathcal{E}|}
    \sum_{e\in\mathcal{E}}
    \frac{S_{m,e}-S_{\mathrm{Random},e}}
    {S_{\mathrm{ANN},e}-S_{\mathrm{Random},e}}
    \times 100\%,
    \label{eq:apr}
\end{equation}
where $\mathcal{E}$ denotes the set of environments and $S_{m,e}$ is the
evaluation score of method $m$ on environment $e$. An APR of $100\%$
corresponds to the ANN baseline (DQN) on average.

\begin{table*}[t]
\centering
\caption{Average maximum return over three random seeds and average performance ratio (APR) on Atari environments after 30M environment transitions.
}
\label{tab:atari}
\setlength{\tabcolsep}{4pt}
\renewcommand{\arraystretch}{1.12}
\resizebox{0.99\textwidth}{!}{%
\begin{tabular}{cccccccccc}
\toprule
Timesteps &  Method & BeamRider & Breakout & Enduro & Pong & Q*bert & Seaquest & SpaceInvaders & APR \\
\midrule
\multirow{2}{*}{--} &   Random & 354.64 & 0.97 & 0.00 & -20.26 & 162.08 & 60.20 & 127.83 & 0.00\% \\
&   DQN(ANN)& 6842.99 & 364.51 & 295.79 & 19.86 & 9224.17 & 1960.44 & 1492.50 & 100.00\% \\
\midrule
\multirow{2}{*}{$T=2$} &  DSQN& \textbf{4279.90} & 276.11 & 280.97 & 17.60 & 1300.21 & 2101.49 & 806.36 & 70.75\% \\
&  CMC-DSQN& 4136.44 & \textbf{382.90} & \textbf{316.87} & \textbf{18.34} & \textbf{4186.33} & \textbf{2358.57} & \textbf{1183.40} & \textbf{87.06\%} \\
\midrule
\multirow{2}{*}{$T=3$}&  DSQN& 4705.58 & 292.41 & 273.83 & 17.23 & 4076.25 & \textbf{2863.83} & 935.96 & 83.32\% \\
&  CMC-DSQN& \textbf{5284.37} & \textbf{398.73} & \textbf{313.55} & \textbf{19.59} & \textbf{8782.83} & 1766.93 & \textbf{1401.10} & \textbf{95.57\%} \\
\midrule
\multirow{2}{*}{$T=4$}&  DSQN& 4676.82 & 318.01 & 266.15 & 16.13 & 4306.67 & \textbf{3589.00} & 883.21 & 88.76\% \\
&  CMC-DSQN& \textbf{5904.69} & \textbf{411.70} & \textbf{311.73} & \textbf{19.98} & \textbf{8941.17} & 2399.73 & \textbf{1491.77} & \textbf{103.45\%} \\
\bottomrule
\end{tabular}%
}
\end{table*}

\begin{table*}[t]
\centering
\caption{
Average maximum return over five random seeds and average performance ratio (APR) on MiniAtar environments after 5M environment transitions, where $\pm$ denotes one standard deviation.}
\label{tab:minatar}
\setlength{\tabcolsep}{4pt}
\renewcommand{\arraystretch}{1.12}
\resizebox{0.99\textwidth}{!}{%
\begin{tabular}{cccccccc}
\toprule
Timesteps & methods & Asterix & Breakout & Freeway & Seaquest & SpaceInvaders & APR \\
\midrule
\multirow{2}{*}{---} & Random & $0.493 \pm 0.902$ & $0.366 \pm 0.618$ & $0.370 \pm 0.586$ & $0.116 \pm 0.427$ & $4.057 \pm 3.199$ & 0.00\% \\
 & DQN (ANN) & $32.22 \pm 4.94$ & $27.76 \pm 4.98$ & $62.58 \pm 0.34$ & $49.70 \pm 8.15$ & $138.34 \pm 3.87$ & 100.00\% \\
\midrule
\multirow{9}{*}{$T=2$} 
 & DSQN-CVT & $1.89 \pm 1.14$ & $4.11 \pm 2.40$ & $19.17 \pm 17.72$ & $1.36 \pm 0.75$ & $3.82 \pm 3.77$ & 10.12\% \\
 & DCGS & $18.36 \pm 3.55$ & $11.64 \pm 2.21$ & $59.22 \pm 0.77$ & $26.17 \pm 2.84$ & $60.56 \pm 20.89$ & 57.33\% \\
 & CRPI & $21.02 \pm 2.36$ & $14.92 \pm 0.89$ & $59.70 \pm 0.64$ & $28.56 \pm 2.44$ & $82.02 \pm 21.38$ & 65.72\% \\
 & DSQN & $14.50 \pm 3.04$ & $23.90 \pm 1.22$ & $56.92 \pm 1.06$ & $22.84 \pm 15.67$ & $100.76 \pm 11.51$ & 67.76\% \\
 & pbLN & $16.20 \pm 1.40$ & $26.43 \pm 2.56$ & $55.57 \pm 0.35$ & $13.87 \pm 2.69$ & $92.70 \pm 1.87$ & 65.43\% \\
 & DATSQN& $14.67 \pm 1.29$ & $23.93 \pm 3.37$ & $58.47 \pm 0.84$ & $2.33 \pm 0.21$ & $99.17 \pm 19.28$ & 59.88\% \\
 & CaRe-BN & $17.46 \pm 1.99$ & $21.38 \pm 2.13$ & $58.04 \pm 0.29$ & $34.76 \pm 5.13$ & $97.42 \pm 4.23$ & 72.46\% \\
 & TP-DSQN & $19.06 \pm 4.40$ & $23.34 \pm 2.54$ & $58.38 \pm 0.77$ & $30.38 \pm 10.84$ & $69.02 \pm 14.37$ & 69.01\% \\
 & CMC-DSQN & $27.30 \pm 4.94$ & $23.50 \pm 5.09$ & $56.74 \pm 1.12$ & $41.06 \pm 14.90$ & $150.92 \pm 36.31$ & \textbf{90.30\%} \\
\midrule
\multirow{9}{*}{$T=4$} 
 & DSQN-CVT & $5.56 \pm 3.22$ & $6.67 \pm 2.07$ & $50.97 \pm 3.63$ & $3.98 \pm 2.01$ & $14.29 \pm 9.35$ & 27.15\% \\
 & DCGS & $21.65 \pm 6.96$ & $13.83 \pm 3.96$ & $60.77 \pm 0.70$ & $33.43 \pm 6.41$ & $78.50 \pm 17.75$ & 67.11\% \\
 & CRPI & $26.42 \pm 3.65$ & $17.87 \pm 2.36$ & $61.28 \pm 0.76$ & $43.06 \pm 8.84$ & $102.76 \pm 17.66$ & 80.72\% \\
& DSQN & $22.04 \pm 5.54$ & $22.96 \pm 4.56$ & $57.04 \pm 1.10$ & $33.10 \pm 6.95$ & $109.50 \pm 15.99$ & 77.31\% \\
 & pbLN & $15.70 \pm 0.36$ & $25.03 \pm 2.23$ & $55.57 \pm 0.67$ & $23.20 \pm 2.46$ & $107.67 \pm 2.64$ & 70.08\% \\
 & DATSQN & $26.96 \pm 3.70$ & $25.06 \pm 1.99$ & $58.76 \pm 3.13$ & $35.96 \pm 14.92$ & $87.38 \pm 11.38$ & 80.35\% \\
 & CaRe-BN & $19.82 \pm 2.98$ & $27.92 \pm 3.20$ & $58.56 \pm 0.70$ & $31.88 \pm 5.39$ & $122.38 \pm 10.21$ & 81.44\% \\
 & TP-DSQN & $28.74 \pm 4.34$ & $26.66 \pm 2.31$ & $60.62 \pm 0.51$ & $45.04 \pm 6.24$ & $105.72 \pm 14.27$ & 89.64\% \\
 & CMC-DSQN & $35.40 \pm 2.36$ & $26.92 \pm 4.34$ & $59.78 \pm 0.37$ & $55.78 \pm 8.78$ & $131.44 \pm 13.50$ & \textbf{101.92\%} \\
\midrule
\multirow{9}{*}{$T=6$} 
& DSQN-CVT & $7.60 \pm 2.82$ & $9.15 \pm 2.72$ & $56.46 \pm 2.09$ & $8.27 \pm 2.56$ & $19.27 \pm 12.49$ & 34.48\% \\
 & DCGS & $23.56 \pm 4.84$ & $15.68 \pm 3.88$ & $60.33 \pm 1.14$ & $38.57 \pm 7.76$ & $89.98 \pm 18.49$ & 73.30\% \\
 & CRPI & $25.65 \pm 3.72$ & $18.41 \pm 2.17$ & $61.08 \pm 0.95$ & $47.63 \pm 10.11$ & $111.13 \pm 16.11$ & 83.66\% \\
& DSQN & $20.98 \pm 1.98$ & $28.50 \pm 2.19$ & $57.04 \pm 0.82$ & $33.98 \pm 10.87$ & $115.14 \pm 11.39$ & 81.88\% \\
 & pbLN & $18.70 \pm 3.42$ & $29.23 \pm 2.49$ & $56.07 \pm 0.25$ & $33.00 \pm 8.51$ & $102.57 \pm 11.42$ & 78.39\% \\
 & DATSQN & $29.20 \pm 3.37$ & $23.24 \pm 2.93$ & $57.00 \pm 6.31$ & $38.56 \pm 12.73$ & $98.04 \pm 16.59$ & 82.51\% \\
 & CaRe-BN & $20.43 \pm 1.63$ & $26.33 \pm 1.22$ & $58.10 \pm 0.53$ & $32.47 \pm 2.76$ & $129.83 \pm 17.06$ & 81.87\% \\
 & TP-DSQN & $33.62 \pm 5.33$ & $24.90 \pm 2.35$ & $61.20 \pm 0.82$ & $46.74 \pm 13.35$ & $115.82 \pm 24.09$ & 93.80\% \\
 & CMC-DSQN & $35.44 \pm 3.63$ & $26.88 \pm 2.75$ & $59.38 \pm 0.50$ & $50.12 \pm 10.59$ & $131.48 \pm 13.94$ & \textbf{99.51\%} \\
\bottomrule
\end{tabular}%
}
\end{table*}
\subsection{Main Results}
\label{Sec:main_results}

As shown in Table~\ref{tab:atari}, CMC-DSQN consistently improves the aggregated
performance of low-timestep DSQNs across the seven Atari games. The improvement
is particularly evident at $T=2$, where CMC-DSQN increases the average
performance ratio (APR) by more than $16$ percentage points over the vanilla
DSQN. The advantage persists as the simulation horizon increases.

Similar improvements are observed on MiniAtar, where we further compare
CMC-DSQN with recent directly trained DSQNs and ANN-to-SNN conversion methods,
as summarized in Table~\ref{tab:minatar} and Fig.~\ref{Fig:intro}. CMC-DSQN
consistently outperforms both categories of baselines across different
simulation timesteps. At $T=2$, CMC-DSQN reaches an APR of $90.30\%$, nearly
$20$ percentage points higher than CaRe-BN~\citep{xu2026care}, the strongest
competing DSQN baseline. 

Overall, CMC-DSQN achieves average normalized performance exceeding the ANN baseline at $T=4$ on both Atari and MiniAtar. These results show that common-mode compensation effectively alleviates the performance degradation caused by reducing the number of SNN simulation timesteps.

\subsection{Ablation Studies}
\label{Sec:ablation_study}
\begin{figure}[t]
\begin{center}
\includegraphics[width=0.99\linewidth]{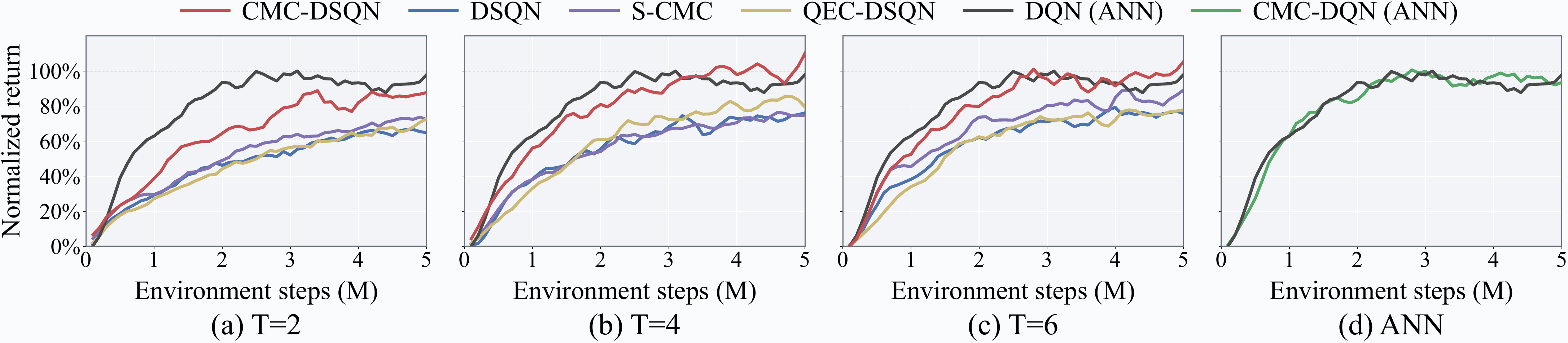}
\caption{
Average normalized return across MiniAtar environments.
(a)--(c) Comparison of different DSQN variants at $T=2,4,6$, respectively.
(d) Comparison of the vanilla ANN DQN and its common-mode compensated variant.
All curves are smoothed uniformly for visual clarity.
}
\label{Fig:ablation}
\end{center}
\end{figure}

\textbf{Is the improvement simply due to additional parameters?} We replace the auxiliary ANN with an SNN, denoted as S-CMC. Figures~\ref{Fig:ablation}(a)--(c) show that S-CMC consistently underperforms CMC-DSQN across different $T$, indicating that the improvement is not due to introducing additional trainable parameters and highlighting the importance of reliable common-mode compensation.

\textbf{Why compensate only the common mode?} We further consider using the auxiliary ANN to compensate the Q-value error rather than only its common-mode component, denoted as QEC-DSQN. Figures~\ref{Fig:ablation}(a)--(c) show that QEC-DSQN improves over the vanilla DSQN but consistently underperforms CMC-DSQN, validating the design choice of compensating only the common-mode errors.

\textbf{Does common-mode compensation also benefit ANN-based DQNs?} We apply the same compensation mechanism to an ANN-based DQN, denoted as CMC-DQN. Figure~\ref{Fig:ablation}(d) shows that CMC-DQN closely matches the vanilla DQN with little additional benefit. This is consistent with Sec.~\ref{Sec:analysis}, where common-mode error dominance is observed in low-timestep SNNs but not in ANNs. This suggests that the improvement of CMC-DSQN primarily addresses an
SNN-specific estimation problem rather than arising from a generic improvement to the underlying RL algorithm.

\section{Conclusion}

We investigated the performance degradation of low-timestep DSQNs from the
perspective of Q-value estimation errors and found that common-mode errors
dominate and affect TD bootstrapping more strongly than differential-mode
errors. Based on this finding, we proposed Common-Mode Compensation Deep Spiking Q-Network (CMC-DSQN), which replaces the SNN
common-mode component with an estimate from an auxiliary ANN. Since the
auxiliary ANN can be removed entirely at inference, CMC-DSQN preserves the
energy-efficiency advantages of SNNs. Experiments on Atari and MiniAtar
environments show that CMC-DSQN substantially outperforms state-of-the-art
DSQN baselines in low-timestep settings and surpasses the ANN baseline at
$T=4$.

\newpage

\bibliography{main}
\bibliographystyle{iclr2027_conference}

\newpage
\appendix

\end{document}

%% file: math_commands.tex
\usepackage{amsmath,amsfonts,bm}

\def\eqref#1{equation~\ref{#1}}
\def\1{\bm{1}}

\DeclareMathAlphabet{\mathsfit}{\encodingdefault}{\sfdefault}{m}{sl}
\SetMathAlphabet{\mathsfit}{bold}{\encodingdefault}{\sfdefault}{bx}{n}